\documentclass[fleqn,10pt,twocolumn]{AROB}
\usepackage[T1]{fontenc}
\usepackage{newtxtext}

\usepackage{amsmath} 
\usepackage[normalem]{ulem}

\usepackage[most]{tcolorbox}
\usepackage{xcolor}
\definecolor{teal}{RGB}{0,128,128}
\usepackage{cite}
\usepackage[colorlinks=true,linkcolor=teal,urlcolor=blue,citecolor=teal]{hyperref}
\tcbuselibrary{skins, breakable, theorems}
\usepackage{cleveref} 
\newcommand{\kara}{} 
\newtcolorbox[auto counter, number within=section, crefname = {Prompt.}{Prompts.}]{prompt}[3][]
{enhanced, breakable = true, fonttitle = \bfseries, fontupper=\scriptsize, left=2mm, right=2mm, top=1mm, bottom=1mm, float, floatplacement=tb,
title = Prompt.~\thetcbcounter.~\if #2\kara \else #2 \fi,
#1,
label = prompt:#3}

\usepackage{booktabs,siunitx,adjustbox}
\usepackage{tabularx}   
\usepackage{booktabs}   
\usepackage{diagbox}    

\title{
Multi-Robot Task Planning for Multi-Object Retrieval Tasks \\with Distributed On-Site Knowledge via Large Language Models
}

\author{%
Kento Murata${}^{1\dagger}$, %
Shoichi Hasegawa${}^{1}$,
Tomochika Ishikawa${}^{1}$,
Yoshinobu Hagiwara${}^{3,4}$,\\%
Akira Taniguchi${}^{2}$, %
Lotfi El Hafi${}^{4}$, %
and Tadahiro Taniguchi${}^{4,5}$%
}

\speaker{Kento Murata}

\affils{
${}^{1}$Graduate School of Information Science and Engineering, Ritsumeikan University, Osaka, Japan\\
(E-mail: \{murata.kento, hasegawa.shoichi, ishikawa.tomochika \}@em.ci.ritsumei.ac.jp)\\
${}^{2}$College of Information Science and Engineering, Ritsumeikan University, Osaka, Japan\\
(E-mail: a.taniguchi@em.ci.ritsumei.ac.jp)\\
${}^{3}$Faculty of Science and Engineering, Soka University, Tokyo, Japan\\
(E-mail: hagiwara@soka-u.jp)\\
${}^{4}$Research Organization of Science and Technology, Ritsumeikan University, Shiga, Japan\\
(E-mail: lotfi.elhafi@em.ci.ritsumei.ac.jp)\\
${}^{5}$Graduate School of Informatics, Kyoto University, Kyoto, Japan\\
(E-mail: taniguchi@i.kyoto-u.ac.jp)\\
}

\abstract{
It is crucial to efficiently execute instructions such as ``Find an apple and a banana.'' or ``Get ready for a field trip,'' which require searching for multiple objects or understanding context-dependent commands.
This study addresses the challenging problem of determining which robot should be assigned to which part of a task when each robot possesses different situational on-site knowledge—specifically, spatial concepts learned from the area designated to it by the user.
We propose a task planning framework that leverages large language models (LLMs) and spatial concepts to decompose natural language instructions into subtasks and allocate them to multiple robots.
We designed a novel few-shot prompting strategy that enables LLMs to infer required objects from ambiguous commands and decompose them into appropriate subtasks.
In our experiments, the proposed method achieved 47/50 successful assignments, outperforming random (28/50) and commonsense-based assignment (26/50).
Furthermore, we conducted qualitative evaluations using two actual mobile manipulators. 
The results demonstrated that our framework could handle instructions—including those involving ad hoc categories such as ``Get ready for a field trip.''—by successfully performing task decomposition, assignment, sequential planning, and execution.
The project website is \url{https://kentomurata0610.github.io/multi-robot-task-planning}.
}


\begin{document}

\maketitle

\section{Introduction}
\label{chap:introduction}
    \begin{figure}[t!]
    \centering
    \includegraphics[width=1\linewidth]{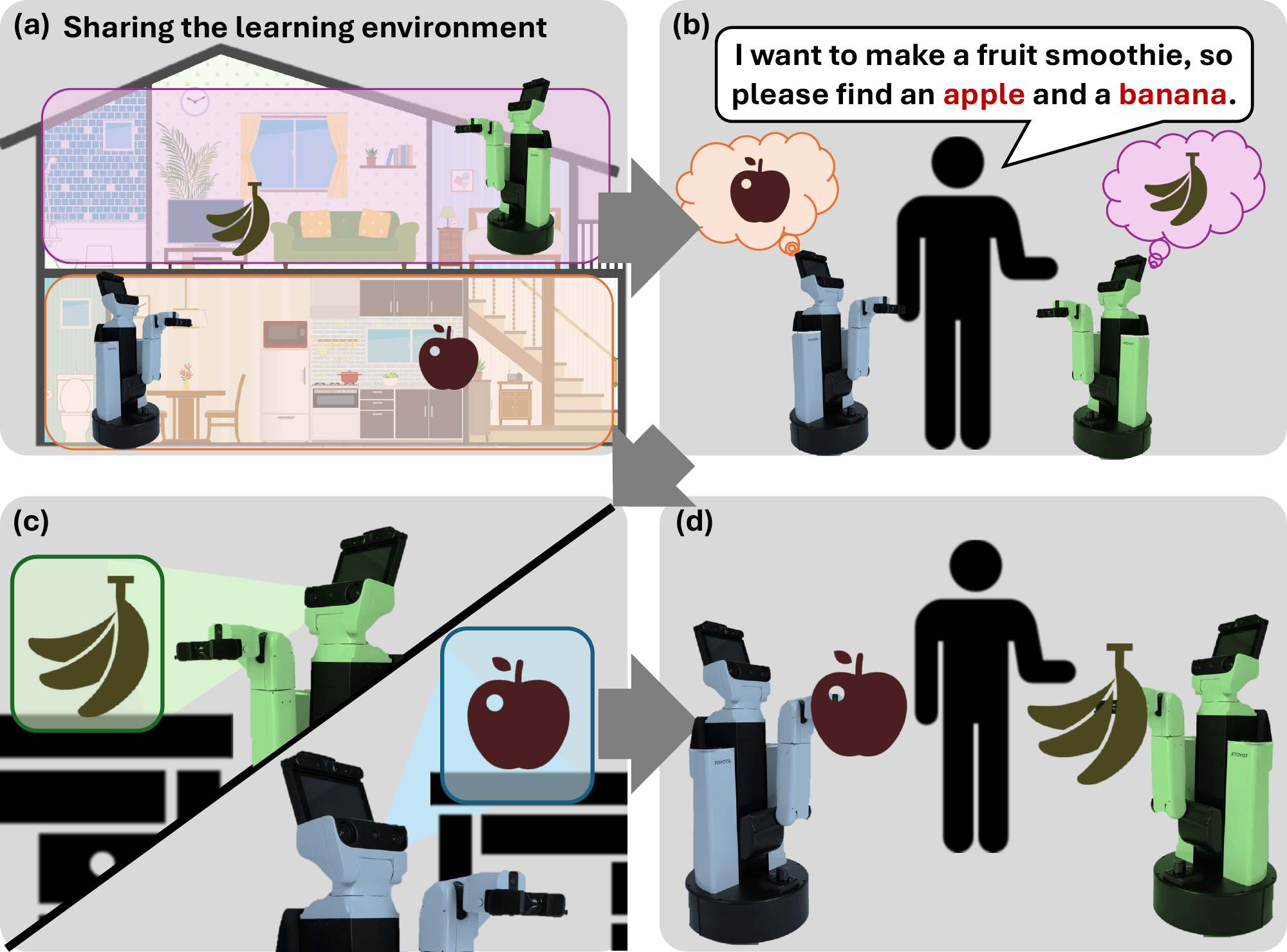}
    \caption{
    Overview of the research problem setting illustrated in four panels (a–d):
    (a) The environment is divided and learned by different robots.
    (b) A user gives a high-level natural language instruction.
    (c) Each robot searches for the requested objects based on its own knowledge.
    (d) The robots return with the objects, completing the instruction.
    For example, two mobile manipulators cooperatively accomplish the instruction ``I want to make a fruit smoothie, so please find an apple and a banana'' in parallel by leveraging their floor-specific knowledge. The performance of this framework is validated through real-world robot experiments described later.
    }
    \label{fig:reserch_overview}
\end{figure}

    This study focuses on cooperative object search by multiple robots in response to natural language instructions within a household environment.  
    In such settings, users often give commands involving multiple target objects, such as: ``I want to make a fruit smoothie, so please find an apple and a banana.''  
    Tasks of this type can be completed more efficiently by distributing subtasks among robots and performing object searches in parallel, compared to using a single robot.  
    Therefore, a system is required that can accurately interpret the given instruction, decompose it into functional subtasks~\cite{kannan2023smart}, and assign them appropriately to each robot.  
    To address this challenge, we propose a multi-robot coordination framework that integrates task decomposition from natural language with task assignment based on each robot's on-site knowledge.



    The central problem of this study is to assign subtasks—produced by task decomposition—to multiple robots so as to maximize the likelihood of success. 
    We make this assignment under the assumption that each robot has distributed, incomplete situational knowledge: its knowledge is biased toward its trained area, and some areas are mutually inaccessible (Fig.~\ref{fig:reserch_overview}(a)(b)).
    This paper investigates the effectiveness of explicitly using spatial concepts—namely, room names and room-wise object presence probabilities—for task assignment.  
    We evaluate this effectiveness using the number of correct assignments as the primary metric.  
    In other words, the focus is on the accuracy of task allocation; other aspects, such as optimization of search time or detailed execution efficiency, are beyond the scope of this paper.  
    For the formal assumptions and problem formulation, refer to Section~\ref{chap:Problem Statement}.

    Large language models (LLMs) have shown strong capabilities in generating plans from natural language inputs~\cite{bommasani2021opportunities,GPT-4}, enabling zero- or few-shot action planning~\cite{hasegawa_rsj23,vemprala2023chatgpt,chen2023open}.  
    Among these, Hasegawa et al.~\cite{hasegawa_rsj23} demonstrated that prompting GPT-4 with spatial information—such as object names and location data—enables not only plan generation but also replanning upon failure and discovery of unsearched objects.

    Multi-robot coordination using LLMs is gaining traction. 
    However, challenges remain regarding the validity of subtask assignments and the scalability of planning~\cite{kannan2023smart,chen2024scalable,mandi2024roco,HasegawaRobotgennba}.  
    While some frameworks propose role negotiation via natural language communication among robots~\cite{zhang2024building}, increasing the number of tasks leads to longer communication sequences and prompts, which in turn raise issues related to token limits and behavioral stability.  
    Kannan et al.~\cite{kannan2023smart} presented a method for decomposing user instructions and assigning subtasks to two robots. 
    However, since their approach does not incorporate on-site knowledge, such as room names or object distributions, held by each robot, the resulting task division may fail to balance the workload or maximize the success probability.





    In this study, we propose a method for decomposing natural language tasks using an LLM and assigning the resulting subtasks to multiple robots based on their local situational knowledge.  
    Each robot then plans and executes actions sequentially, according to its assigned subtask and the spatial knowledge it has learned.  
    An overview of our framework is shown in Fig.~\ref{fig:reserch_overview}.  
    In the illustrated example, the user's instruction ``I want to make a fruit smoothie, so please find an apple and a banana.'' is first decomposed into subtasks by the LLM.  
    The subtasks are then allocated to individual robots based on their knowledge of object locations, allowing them to collaborate efficiently to complete the task.  
    The effectiveness of the proposed framework is evaluated using real mobile manipulators in a RoboCup environment, which simulates a domestic setting (see demonstration video~\footnote{https://youtu.be/LMoWAp\_kPhk}).

    In addition, we adopt the spatial concept model~\cite{taniguchi2020improved,hasegawa2023integrating} as the basis for on-site knowledge used in multi-robot task assignment.  
    This model can estimate both place names and room-wise object presence probabilities in an unsupervised manner, utilizing observations, including images and self-localization data.  
    We utilize these spatial inferences—based on coherent spatial units—as core reasoning cues for task decomposition and robot-to-subtask assignment decisions.

    The same framework has previously been evaluated from the perspective of learning cost~\cite{HasegawaRobotgennba}.  
    In that work, two mobile manipulators were deployed across two floors in a simulated home environment, and the number of learning episodes for object-location associations (ranging from 0 to 60) was varied in simulation.  
    The object search task was evaluated using the number of successful discoveries per room visit as the success score.  
    In environments with common-sense object placements, the proposed method exceeded the satisfaction threshold (13.5 out of 15 commands) even with only 0 or 10 learning episodes, and it reduced the required observation data by more than eightfold compared to baseline conditions.
    In environments with unique object layouts, the method achieved the satisfaction threshold with 60 learning episodes, reducing data requirements by more than half.  
    These results demonstrate the effectiveness of integrating spatial concept models and LLM-based inference, as well as the utility of LLMs for task assignment.  
    In this paper, we build on this framework and focus on validating its task allocation performance using real robots in a real-world setting.

    The main contributions of this work are as follows:
    
    \begin{itemize}
        \item This work shows that practical multi-robot coordination can be achieved by integrating LLM-based task decomposition with subtask assignment grounded in spatial concepts.
        \item This work shows that explicitly encoding room names and room-wise object distributions—as estimated by the spatial concept model—enables subtask allocations that align with each robot's on-site knowledge.
        \item Our experiments with two mobile manipulators in a real RoboCup environment show the practical utility of the proposed system in executing ambiguous instructions.
    \end{itemize}

    The remainder of this paper is organized as follows.
    Section~\ref{chap:Problem Statement} defines the multi-robot object search problem in a household context.
    Section~\ref{chap:preparation} reviews related work on LLMs, multi-robot planning, and spatial concept models.
    Section~\ref{chap:method} describes the proposed framework.
    Section~\ref{chap:Experiment I} reports baseline comparisons, and Section~\ref{chap:Experiment II} presents real-robot experiments in the RoboCup environment.
    Finally, Section~\ref{chap:conclusion} concludes with findings and future directions.

\section{Problem Statement}
\label{chap:Problem Statement}

In this section, we define the problem of enabling multiple robots to collaboratively execute tasks based on a user's natural language instructions in a household environment.  
We also identify three key challenges required for successful execution: (i) distributed and incomplete situational knowledge, (ii) ambiguous and diverse instructions, and (iii) sequential planning with feedback.  
This problem assumes an integrated process in which language instructions are decomposed into executable subtasks, allocated based on distributed on-site knowledge, and smoothly connected to sequential planning and execution~\cite{feng2020overview,bommasani2021opportunities,GPT-4,chen2023open,vemprala2023chatgpt}.

\subsection{Distributed and Incomplete On-Site Knowledge with Limited Search Coverage}
\label{chap:Distributed and incomplete field knowledge and limited search coverage}

In the target environments of this study, each robot has access to different areas for learning and operation.  
As a result, on-site knowledge—such as room names and room-wise object presence probabilities—is distributed and incomplete.  
This issue is evident in scenarios involving floor-wise task division or mutually inaccessible regions, where robots cannot uniformly evaluate the entire environment.  
In such cases, it becomes essential to estimate which robot should be assigned to which subtask to maximize the likelihood of success, considering both knowledge asymmetry and access limitations~\cite{feng2020overview}.  
We address this challenge by representing on-site knowledge using a spatial concept model, and we base assignment decisions on the inferred room names and object distributions from observations~\cite{taniguchi2020improved,hasegawa2023integrating}.

\subsection{Natural Language Instructions with Ambiguity and Diversity}
\label{chap:Natural language instructions with ambiguity and diversity}

Natural language commands in domestic settings often include ambiguous expressions, such as instructions to search for multiple objects or references to necessary items that are not explicitly stated.  
These instructions require commonsense-based completion and semantic decomposition into subtasks.  
To handle such ambiguity, the system must accurately interpret user commands, convert them into appropriate subtasks, and allocate them in a manner consistent with each robot's on-site knowledge~\cite{kannan2023smart}.






\section{Related Work}
\label{chap:preparation}

\subsection{LLM-based Action Planning for Robot Systems}
\label{sec:Related works_planning_by_LLM}

In single-robot systems, LLMs combined with feasibility estimation have shown promise for turning language into executable action plans. SayCan~\cite{ahn2022can} maps natural-language commands to skill sequences but lacks robust mechanisms for replanning under failures or environmental changes, limiting reliability. Hence, effective systems should integrate situational knowledge and closed-loop, sequential feedback to update plans dynamically.


A promising direction is to tightly couple a spatial concept model, an LLM, and a behavior execution engine. Hasegawa et al. integrate structured environmental knowledge with GPT-4 and execute via FlexBE~\cite{FlexBE} in a closed loop using succeed/failed feedback~\cite{hasegawa_rsj23}, but validation is limited to single-robot settings. We extend this line by enabling multi-robot coordination through LLM-based instruction decomposition and spatial-concept–grounded subtask allocation.

\subsection{LLM-based Planning for Multi-Robot Coordination}
\label{sec:Related works_multi_robot_by_LLM}

LLM-based multi-robot planning is rapidly advancing~\cite{kannan2023smart,chen2024scalable,mandi2024roco}. 
Existing methods fall into two paradigms: (i) dialogue-based coordination, where robots negotiate plans in natural language, and (ii) centralized allocation, where a parent LLM decomposes the task and assigns subtasks. 
The former enables shared understanding, whereas the latter reduces interaction overhead by centrally handling decomposition and dependencies.


In dialogue-based coordination, robots iteratively refine plans via natural-language exchanges, but this paradigm suffers from scalability limits as prompts grow and context windows are exceeded. 
Zhang et al.~\cite{zhang2024building} share goals, dialogue history, and task progress in prompts and use Zero-shot CoT~\cite{kojima2022large} (``Let's think step by step'') to encourage reasoning, yet longer dialogues increase token pressure and destabilize responses. 
To avoid this bottleneck, we adopt centralized allocation, first decomposing tasks systematically and then assigning subtasks to robots.


In centralized allocation, a parent LLM is responsible for task decomposition, subtask assignment, and sequencing, thereby mitigating the scaling issues associated with dialogue-based approaches.  
SMART-LLM~\cite{kannan2023smart} decomposes complex tasks into subtasks and assigns them in order, based on skill differences and task dependencies.  
It also incorporates Chain-of-Thought prompting~\cite{wei2022chain} through systematic examples to promote intermediate reasoning and improve consistency in decomposition and assignment.  
However, SMART-LLM does not explicitly incorporate spatial semantics such as object placement or room names, relying instead on fixed coordinates. This limits generalizability and feasibility in dynamic or unknown environments.  
Our study addresses this limitation by integrating spatial concepts (room names and room-wise object presence probabilities) to improve the semantic grounding of task decomposition and allocation.

Centralized allocation has been explored in various frameworks.
COHERENT~\cite{liu2024coherent} introduces a hierarchical design that separates high-level assignment from low-level execution through a Proposal–Execution–Feedback–Adjustment loop, achieving robustness in long-horizon tasks. 
AutoHMA-LLM~\cite{yang2025autohma} adopts a hybrid cloud–edge architecture where cloud-based LLMs perform high-level scheduling while lightweight on-device agents handle feasibility checks and execution, enhancing both efficiency and robustness. 
El Hafi et al.~\cite{el2025public} further bridge decomposition, allocation, and execution by encoding preconditions, dependencies, and resource constraints into prompts. 
Building on these insights, our study strengthens centralized allocation by embedding spatial concepts—room names and room-wise object presence probabilities learned within each robot's designated area—into task decomposition and subtask assignment, thereby grounding allocations in on-site knowledge while maintaining scalability.
This study positions itself as a foundational exploration of integrating environment-dependent spatial concepts into language-based task planning with a focus on reproducible evaluation. While related work has reported live demonstrations with heterogeneous robots~\cite{el2025public}, our emphasis is on measuring the effectiveness of planning and allocation under controlled settings.

Based on the above discussions, we conclude that integrating spatial semantics and feasibility-aware feedback into centralized allocation is a promising direction for achieving both scalability and robustness.  
In single-robot settings, integrating spatial knowledge and feedback has been shown to improve task efficiency~\cite{hasegawa_rsj23}.  
Dialogue-based coordination offers flexibility but suffers from prompt bloating and scaling issues~\cite{zhang2024building,kojima2022large}.  
Centralized allocation mitigates these issues but often lacks spatial grounding~\cite{kannan2023smart}, while hierarchical separation~\cite{liu2024coherent} and cloud–edge delegation~\cite{yang2025autohma} have proven effective for long-horizon planning and communication efficiency.  
The utility of LLM-based task decomposition itself has also been confirmed in real-world scenarios~\cite{el2025public}.  

Building on these insights, this study proposes the following:  
(i) embedding spatial concepts (room names and room-wise object presence probabilities) into prompt design and task allocation objectives;  
(ii) introducing lightweight feedback mechanisms that validate action feasibility and propagate failure reasons back to the central model; 
(iii) designing a centralized allocation pipeline that ensures stability while minimizing communication and token costs.

\section{Proposed Method}
\label{chap:method}
\begin{figure*}[tb]
    \centering
    \includegraphics[width=0.85\linewidth]{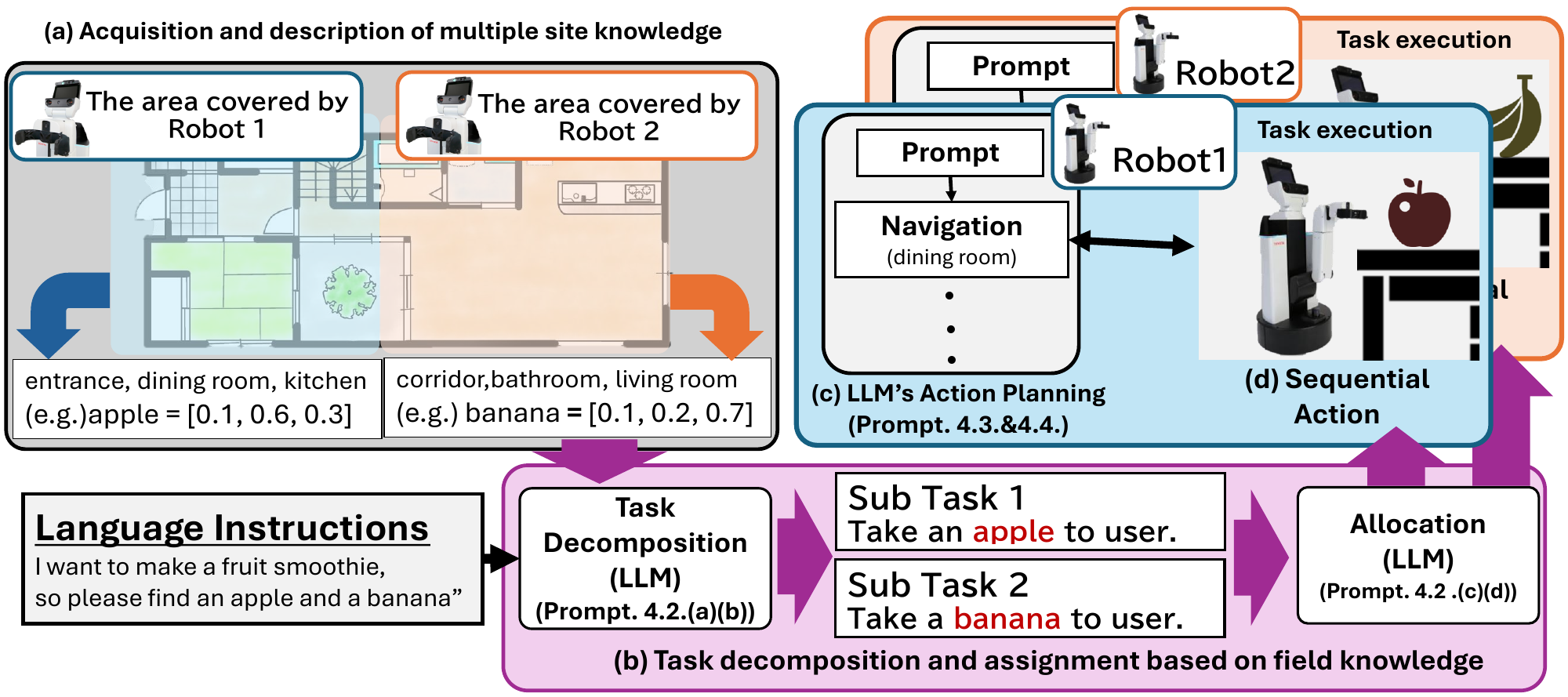}
    \caption{Overview of the proposed pipeline: (a) Each robot organizes its on-site knowledge—room names, vocabulary, and object occurrence probabilities—using the spatial concept model; (b) GPT-4 uses this knowledge as prompt components to perform subtask decomposition and allocation; (c) action-planning prompts are generated accordingly; and (d) FlexBE executes the actions via a state machine and returns feedback.}
    \label{fig:method_overview}
\end{figure*}

We integrate an LLM (GPT-4) with a spatial concept model that links places (e.g., kitchen) with likely objects. Using this on-site knowledge, our four-stage pipeline (Fig.~\ref{fig:method_overview}) performs decomposition, allocation, sequential planning, and execution to maximize each robot’s success probability.
This framework aims to assign roles that maximize the success probability for each robot and maintain robustness in novel or complex environments.


\subsection{Acquisition and Representation of On-site Knowledge via Spatial Concepts}
\label{sec:method_location}

This section describes how the spatial concept model probabilistically represents ``where what is likely to be'' in the environment, and how this information is transformed into a prompt that GPT-4 can use.  
The spatial concept model is a probabilistic framework that links map regions with (i) place names as referred to by users (e.g., kitchen, bedroom), (ii) visual features from sensor observations, and (iii) location distributions.  
Through interaction with the environment, the robot learns spatial concepts that associate linguistic and perceptual information~\cite{hasegawa2023integrating}.

In our method, we construct the following two knowledge components for each region:  
1) a set of frequently observed place-related vocabulary, and  
2) an object-to-location probability table.  
These are formatted as prompt components (a place vocabulary list and an object–location probability table) and passed to GPT-4 to inform its decisions during task decomposition and allocation.

\begin{figure}[tb]
    \centering
    \includegraphics[width=1\linewidth]{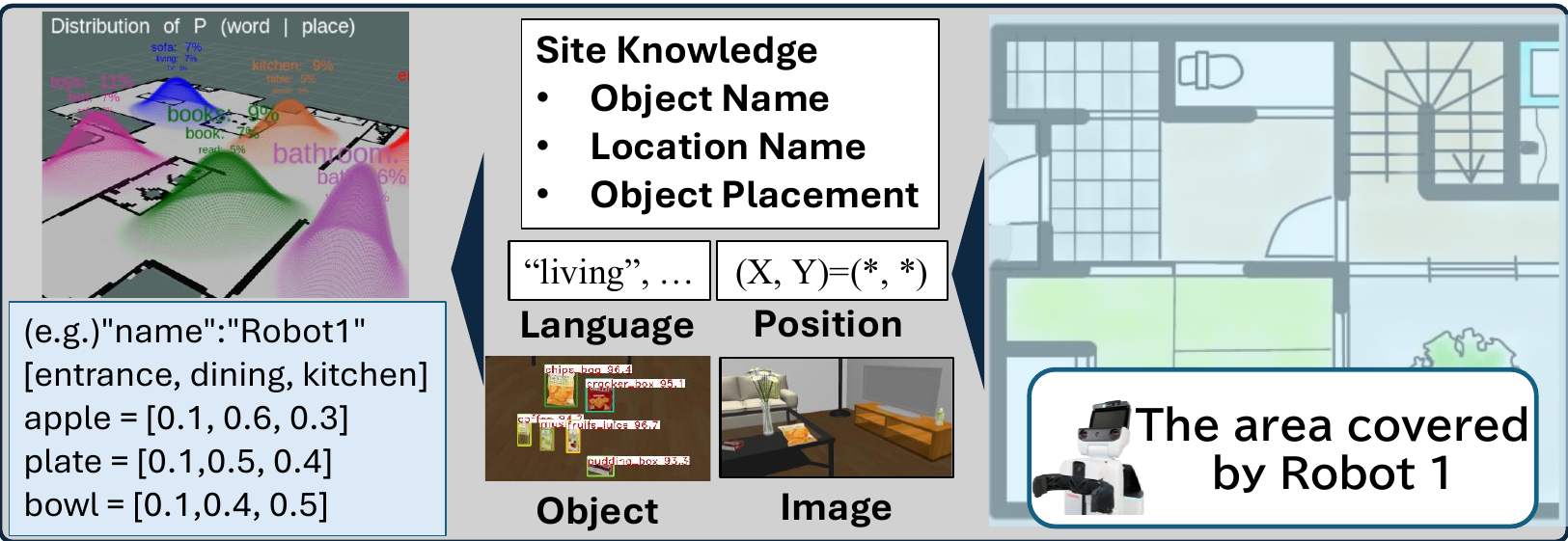}
    \caption{Illustration of how the robot infers both the frequently observed place-related vocabulary and the object–location existence probabilities in each area based on sensor observations.}
    \label{fig:a_SpCo}
\end{figure}

First, the system extracts likely place names for each area and formats them into prompts interpretable by GPT-4.  
As illustrated in Fig.~\ref{fig:a_SpCo}, the resulting place-vocabulary list and object–location probability table are then converted into prompts to support downstream task decomposition, allocation, and planning.
Specifically:  
(i) During the learning phase, the robot collects user utterances and extracts candidate place vocabulary (e.g., ``kitchen,'' ``bedroom'').  
(ii) Based on the spatial concept model, the occurrence probability of word $w_t$ in map region $i_t$ is estimated using Eq.~\ref{eq:cross_modal_place}, which intuitively reflects how likely a given word is to appear in that region.  
\begin{equation}
    \label{eq:cross_modal_place}
    \begin{aligned}[b]
    &P(w_t \mid i_t, W, \phi, \pi) \\
    &= \sum_{C_t} {\mathrm{Mult}}(w_t \mid W_{C_t}) 
    {\mathrm{Mult}}(i_t \mid \phi_{C_t}) 
    {\mathrm{Mult}}(C_t \mid \pi)
    \end{aligned}
\end{equation}
Here, $\mathrm{Mult}(w_t \mid W_{C_t})$ denotes the categorical (multinomial) probability of generating word $w_t$ from the vocabulary distribution $W_{C_t}$ associated with concept $C_t$.

These estimates are derived from parameters of categorical or multinomial distributions (e.g., $\pi$, $\phi_l$, $W_l$) and yield a region-specific set of frequently observed place-related vocabulary.  
(iii) All high-probability terms for each region are then listed and embedded into prompts in the format shown in \cref{prompt:place_vocab}.  
This vocabulary is statistically grounded in real-world data, which limits GPT-4's output to contextually relevant language and reduces hallucinations during inference.  
Moreover, the structured linguistic cues help improve the quality of spatial reasoning and subtask decomposition~\cite{kannan2023smart}.  
Prior work has demonstrated that integrating multimodal spatial concepts with probabilistic inference enables stable learning and usage of place-related vocabulary, even with limited training samples~\cite{hasegawa2023integrating}.

    \begin{prompt}{Location-Related Vocabulary (home environment). }{place_vocab}
        1: There are three location areas in a home environment.\\
        2: Your initial position is outside of the three rooms.\\
        3: In each region, words related to the following locations are likely to be observed.\\
        4: place1: [living\_room, sofa, desk, chair, tv]\\
        5: place2: [sink, refrigerator, desk, chair, kitchen]\\
        6: place3: [toy, shelf, toy\_room, box, bed]
    \end{prompt}

Next, we estimate the probability of where each object is likely to be found and embed this information into a prompt as a probability table that GPT-4 can directly utilize.  
During the learning phase, the robot collects object labels detected via an object detector and uses the spatial concept model to compute the likelihood of observing each object $o_t$ in each region $i_t$, using Eq.~\ref{eq:cross_modal_object}.  
\begin{equation}
    \label{eq:cross_modal_object}
    \begin{aligned}[b]
    &P(i_t \mid o_t, \xi, \phi, \pi) \\
    &= \sum_{C_t} {\mathrm{Mult}}(i_t \mid \phi_{C_t}) 
    {\mathrm{Mult}}(o_t \mid \xi_{C_t}) {\mathrm{Mult}}(C_t \mid \pi)
    \end{aligned}
\end{equation}

Intuitively, this corresponds to a statistically normalized measure of how often the object was observed in a given room, and is represented by multinomial distribution parameters (denoted as $\xi_l$ in the main text).  
The results are formatted as a table, where each row corresponds to an object and each column indicates the presence probability for a specific location.  
This table is embedded into the prompt following the format shown in \cref{prompt:loc_infer_prompt}:(c), enabling the LLM to prioritize candidate locations and delivery paths based on probabilistic reasoning.

This probability table is reused during downstream task decomposition and allocation steps as a shared basis for decision-making.  
It provides a consistent foundation for determining which robot is most likely to succeed in executing each subtask.  
Related work has shown that spatial–object relationships inferred via probabilistic models can significantly improve search efficiency~\cite{hasegawa2023integrating}, and the proposed design aligns with these findings.


\subsection{Task Decomposition and Allocation Based on On-site Knowledge}
\label{sec:method_task_decomposition}
Using on-site knowledge (place vocabulary and object–location probabilities), GPT-4 performs both task decomposition and allocation in two prompts, assigning each subtask to the robot with the higher expected success (Fig.~\ref{fig:method_overview}(b)).


For task decomposition, each instruction is broken down into minimal executable units that can be carried out in parallel by different robots.  
We adapted the SMART-LLM prompt by removing mass-related object descriptions and instead aligning the subtasks with the skill set available on the robots: \texttt{Navigation}, \texttt{Object Detection}, \texttt{Pick}, and \texttt{Place} (\cref{prompt:loc_infer_prompt}:(a), (b)).  
In the subsequent allocation step, we append each robot's object-location probability list to the prompt and instruct GPT-4 to infer which robot is likely to complete each subtask successfully, based on probabilistic justifications.  
Assignments are then determined according to GPT-4's reasoning, as shown in \cref{prompt:loc_infer_prompt}:(c), (d).

This design enables execution plans that reflect environment-specific prior knowledge, reducing redundant exploration and role conflicts.  
Consequently, the overall task completion time and reliability of the multi-robot system are expected to improve.



    \begin{prompt}{Prompt from Location Concept Model Inference. }{loc_infer_prompt}
        1: \textbf{(a) Skills}\\
        2: navigation, object\_detection, pick, put\\
        3: \\
        4: \textbf{(b) Objects in the environment}\\
        5: plate, bowl, pitcher\_base, banana, apple, orange, cracker\_box, pudding\_box, chips\_bag, coffee, ...\\
        6: \\
        7: \textbf{(c) room-wise object presence probabilities observed by robots}\\
        8: Robot1\\
        9: ``List of probabilities that an object exists'':\\
        10: [entrance, dining, living\_room, office\_room, kitchen]\\
        11: pitcher\_base = [0.136, 0.848, 0.004, 0.006, 0.006]\\
        12: bowl = [0.136, 0.848, 0.004, 0.006, 0.006]\\
        13: ----------------\\
        14: Robot2\\
        15: ``List of probabilities that an object exists'':\\
        16: [front\_of\_stairs, corridor, bathroom, child\_room, parent\_room]\\
        17: banana = [0.1083, 0.0822, 0.180, 0.229, 0.327]\\
        18: cracker\_box = [0.0, 0.10109009418, 0.2, 0.2, 0.2980]\\
        19: \\
        20: \textbf{(d) Rule for assigning subtasks}\\
        21: \# IMPORTANT: Subtasks are assigned taking into consideration the objects listed in the ``List of probabilities that an object exists'' that each robot has.
    \end{prompt}

\begin{itemize}
    \item We added a table of room-wise object presence probabilities inferred by the spatial concept model to the GPT-4 prompt.
    \item We improved the assignment process by enabling GPT-4 to predict which robot is more likely to complete each subtask successfully based on these probabilities, and to assign tasks accordingly.
\end{itemize}


\subsection{Sequential Action Planning Based on On-Site Knowledge}
\label{sec:method_plannig}

For sequential action planning grounded in on-site knowledge, we provide GPT-4 with the following information in a single prompt: the assigned subtask, place-related vocabulary, room-wise object presence probabilities, the robot's skill set, and illustrative examples.  
Based on this input, GPT-4 generates a step-by-step plan from object search to grasping, transport, and placement (see Fig.~\ref{fig:cd_planning}, \cref{prompt:robot_behaviors}, \cref{prompt:task_dialogue}).

Specifically, the prompt is constructed to follow these steps:  
(1) Clearly state the subtask goal and its preconditions (e.g., target location, object name, safety constraints).  
(2) Use room-wise object presence probabilities from the spatial concept model to prioritize which locations should be searched first.  
(3) Map each subgoal to the appropriate skill from the available set: \texttt{Navigation}, \texttt{Object Detection}, \texttt{Pick}, and \texttt{Place}.  
(4) Define input/output conditions and termination criteria (success/failure) for each step.  
(5) Provide fallback options for failure cases—such as alternative search locations or re-detection—based on the underlying probabilities.

This step-by-step prompt design allows GPT-4 to generate executable action sequences that reflect environment-specific prior knowledge.  
By grounding each decision in a probabilistic context, the system can avoid redundant movements and repeated search cycles.  
As a result, the robot is more likely to converge on a successful task execution path, even in previously unseen environments.



    \begin{prompt}{List of Behaviors (robot skill set).}{robot_behaviors}
        1: navigation (location\_name): move to location\_name\\
        2: object\_detection (object\_name): detect an object\_name and its position from a captured image\\
        3: pick (object\_name): pick up an object\_name\\
        4: place (location\_name): place an object to the location\_name\\
        5: \\
        6: These behaviors return ``succeeded'' or ``failed''. If ``failed'' is returned, try the same or another behavior again.\\
        7: Do not ask back anything about the user's instructions.
    \end{prompt}

    \begin{prompt}{Example Dialogue for Solving a Task (user instructions).}{task_dialogue}
        1: USER : bring the cup to the kitchen\\
        2: ASSISTANT : navigation (living\_room)\\
        3: USER : succeeded\\
        4: ASSISTANT : object\_detection (cup)\\
        5: USER : succeeded\\
        6: ASSISTANT : pick (cup)\\
        7: USER : failed\\
        8: ASSISTANT : pick (cup)\\
        9: USER : succeeded\\
        10: ASSISTANT : navigation (kitchen)\\
        11: USER : succeeded\%finished\\
        12: USER : ``I need you to locate a cup for me.''
    \end{prompt}

\subsection{Robot Execution and Feedback via FlexBE}
\label{sec:method_FlexBE}

The action plans generated by GPT-4 are executed safely using FlexBE~\cite{FlexBE}, which operates as a state machine. Each execution outcome is immediately fed back to inform subsequent decision-making.  
Specifically, the skills described in natural language by GPT-4—\texttt{Navigation}, \texttt{Object Detection}, \texttt{Pick}, and \texttt{Place}—are implemented as FlexBE ``states'' following the design~\cite{hasegawa_rsj23}.  
Each state monitors sensor conditions while executing low-level behaviors and returns \texttt{succeed} if the termination condition is met, or \texttt{fail} otherwise.

During execution, the system operates in a closed loop:  
(i) the next state is triggered according to the assigned subtask;  
(ii) the result label (\texttt{succeed}/\texttt{fail}) and contextual observations (e.g., undetected object, blocked path, grasp failure) are returned to GPT-4 as input;  
(iii) GPT-4 uses this feedback, along with spatial knowledge (place vocabulary and room-wise object presence probabilities), to replan or select fallback actions—such as searching in an alternative location, rerouting, or relaxing grasping constraints.

\begin{figure}[tb]
    \centering
    \includegraphics[width=1\linewidth]{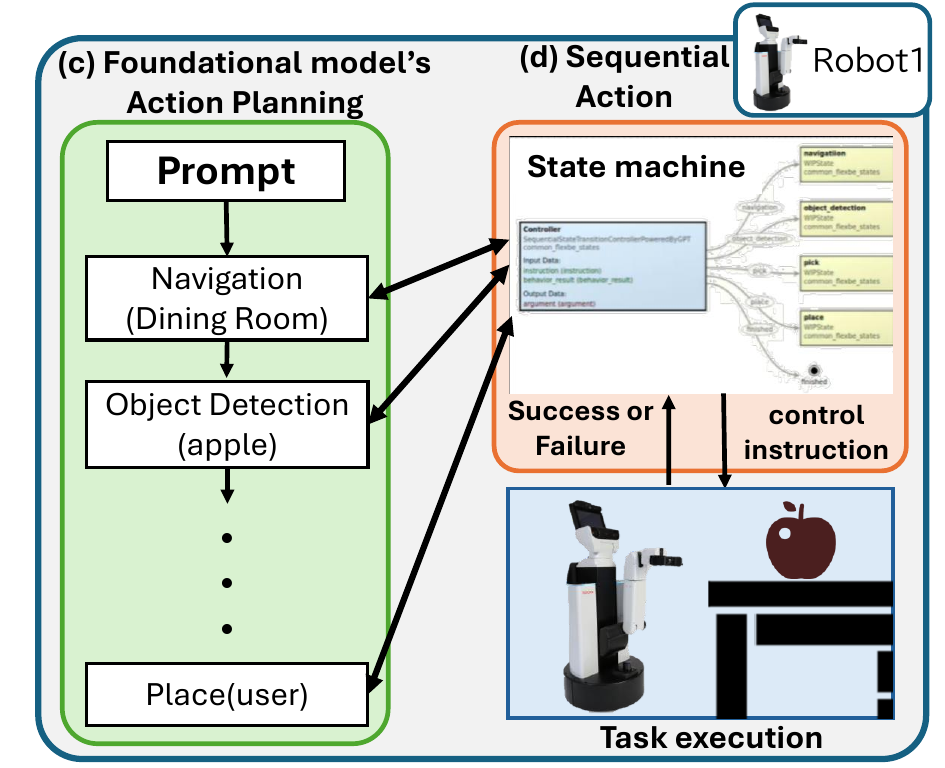}
    \caption{Execution–replanning loop: Skills generated by GPT-4 are executed as FlexBE state machines. Each execution result and observation (e.g., an undetected object, a blocked path, or a grasp failure) is immediately fed back to GPT-4, which replans accordingly. Since each robot operates its own loop independently, parallel task execution is possible based on prior allocation decisions.}
    \label{fig:cd_planning}
\end{figure}

This execution–replanning loop ensures consistency between the plan and real-world behavior, and enables robust recovery from unexpected failures in unfamiliar environments (see Fig.~\ref{fig:cd_planning}).  
Since each robot runs an independent state machine based on its assigned role, the system can progress in parallel, improving overall task completion time and success rate.

\subsection{Handling Ambiguous Instructions}
\label{sec:Ambiguous_language_instructions}

To handle ambiguous natural language instructions, we integrate GPT-4's commonsense reasoning with spatial knowledge from the concept model to perform a unified process:  
(i) hypothesize necessary objects, (ii) convert them into subtasks, and (iii) allocate them to appropriate robots.

In ambiguous instructions, the target objects are often not explicitly specified. We therefore first use GPT-4’s commonsense reasoning, guided by a few-shot prompting design, to infer a concise set of required items sufficient to fulfill the command (e.g., ``Get ready for a field trip.'' → ``water bottle'', ``backpack''), before converting them into executable subtasks and allocating them to robots. 
To suppress over- or under-generation, the candidate set is anchored to known object labels (and synonyms), and search orders are constrained to follow locations with higher object-presence probabilities from the spatial concept model. 
This design enables robots to start in parallel from the highest-probability locations.
A qualitative demonstration in the RoboCup environment corroborates this behavior.

First, we extend the SMART-LLM~\cite{kannan2023smart} prompting strategy to support ambiguity resolution.  
We design a few-shot examples (e.g., ``Get ready for a field trip.'' -> water bottle, backpack) that serve as inference templates, prompting GPT-4 to infer possible target items and generate corresponding action sequences.

Next, for each hypothesized item, we reference room-wise object presence probabilities and associated place vocabulary (e.g., kitchen, bedroom) from the spatial concept model.  
GPT-4 estimates which robot is most likely to succeed in retrieving each item based on location knowledge and assigns the tasks accordingly (\cref{prompt:task_decomposition}).

In the prompt design, we explicitly constrain the search space by anchoring candidate objects to the set of known object labels (and their synonyms), and by specifying that search orders should follow locations with higher presence probabilities.  
These constraints suppress over-generation and guide GPT-4 toward generating rational, context-aware plans grounded in environmental knowledge.

As a result, each robot can begin acting in parallel from the most likely search locations, leading to reduced preparation time and improved task success rate compared to single-robot execution.

    \begin{prompt}{Task Decomposition Example (prompt for ambiguous instruction).}{task_decomposition}
        1: \textbf{Task Description:}\\
        2: Prepare for the excursion.\\
        3: \\
        4: \textbf{SubTask 1:} Bring a water bottle.\\
        5: \textbf{SubTask 2:} Bring a backpack.
    \end{prompt}

\section{Experiment~1}
\label{chap:Experiment I}

The objective of this experiment is to verify whether subtasks obtained via task decomposition can be correctly assigned to the appropriate robots, based on the on-site knowledge each robot has acquired through its own learning.  
Specifically, we evaluate whether each robot can utilize its domain-specific knowledge (e.g., room names and object-location likelihoods) to influence allocation decisions and whether the resulting assignments align with the robots that actually possess that knowledge.

\subsection{Experimental Setup}
\label{sec:Experiment_term}

\begin{figure}[tbh]
    \centering
    \includegraphics[width=1\linewidth]{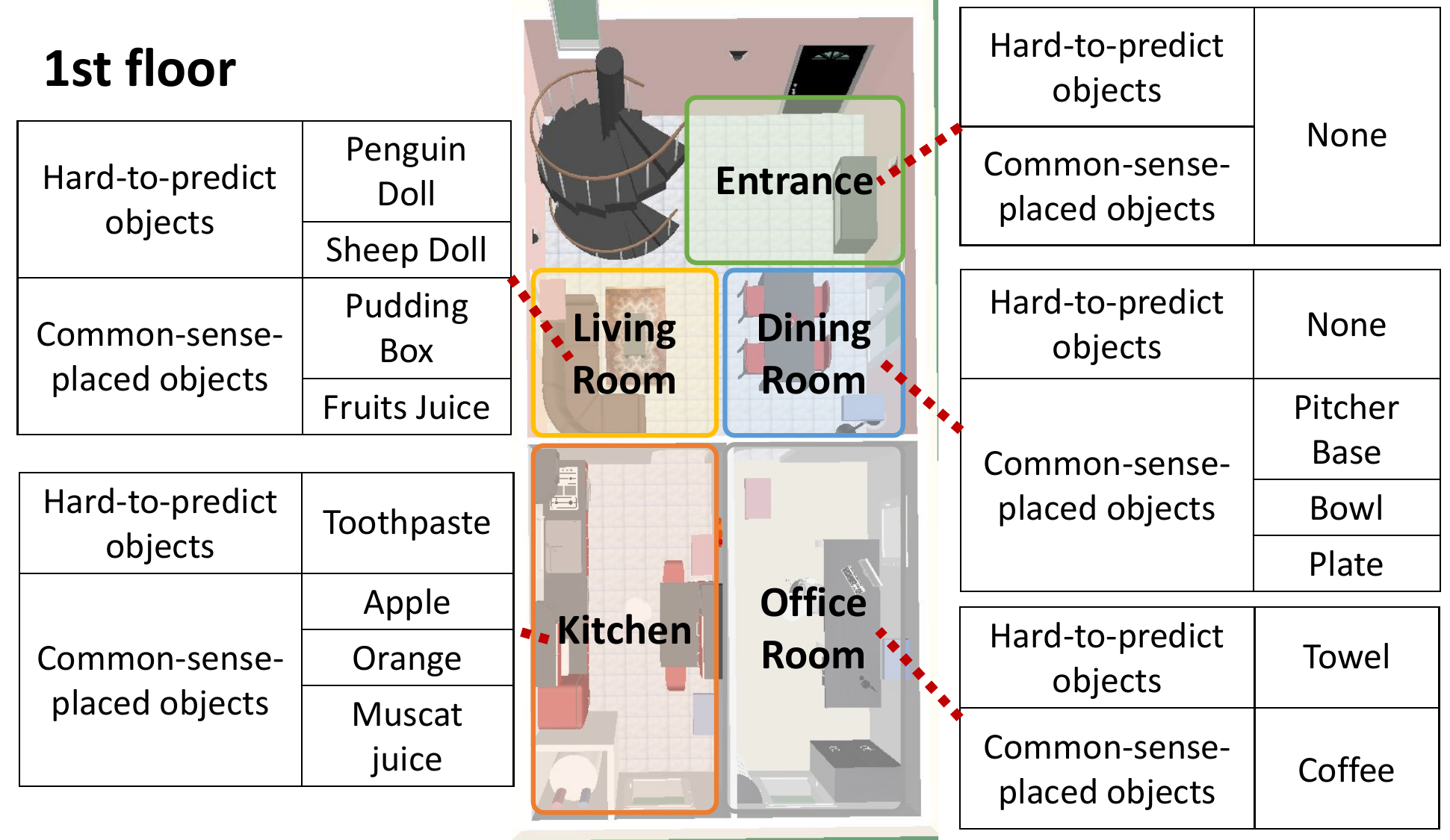}
    \includegraphics[width=1\linewidth]{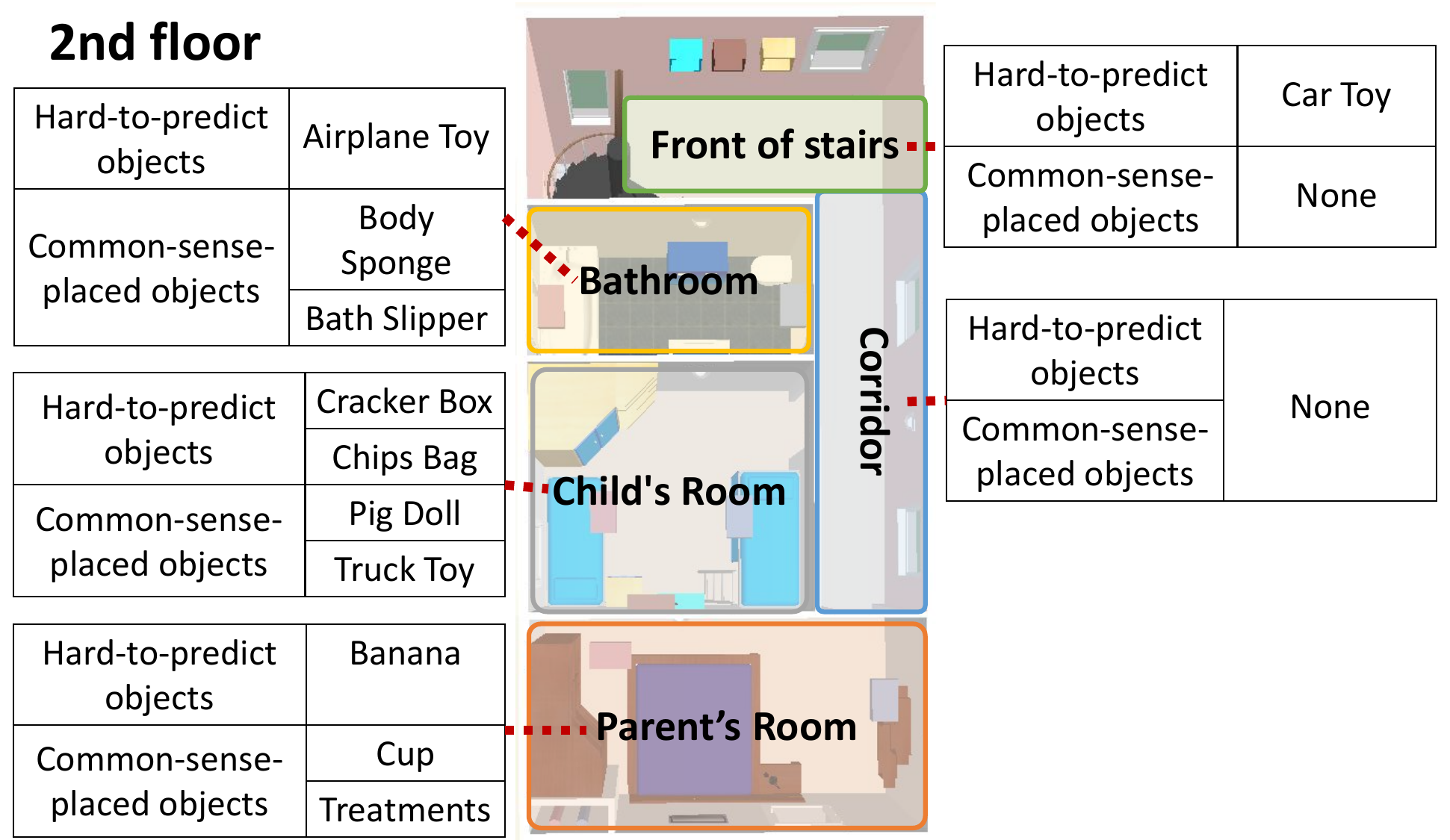}
    \caption{Object placement in the evaluation environment created using Sweet Home 3D. The environment consists of 10 rooms (1F: Living Room, Dining Room, Entrance, Kitchen, Office Room; 2F: Bathroom, Child’s Room, Parent Room, Front of Stairs, Corridor). 24 objects were placed. Objects labeled as ``common-sense placed'' were deemed appropriate for their rooms by GPT-4, while ``hard-to-predict objects'' require spatial knowledge for accurate localization.}
    \label{fig:object_placement}
\end{figure}

As shown in Fig.~\ref{fig:object_placement}, we constructed a simulated two-floor domestic environment containing 10 rooms (five per floor) using Sweet Home 3D\footnote{https://www.sweethome3d.com/ja/gallery.jsp}.  
The first floor consists of a Living Room, Dining Room, Entrance, Kitchen, and Office Room, while the second floor includes a Bathroom, Child's Room, Parents' Room, Front of Stairs, and Corridor.  
We imposed a physical separation between floors such that robots cannot move between them, simulating a scenario where a wide indoor space is divided and must be collaboratively covered by two robots, each confined to its assigned area.

Unlike single-robot full-coverage exploration, each robot independently acquires spatial knowledge—including room names, object names, and semantic maps—within its own floor and performs task execution based on that on-site knowledge.

To systematically control task diversity and allocation difficulty, we placed a total of 24 objects across the two floors. These were classified into two categories:  
\textbf{Hard-to-predict objects}, for which inference based on acquired spatial concepts is essential (i.e., common sense alone is insufficient), and  
\textbf{Common-sense-placed objects}, whose likely locations can be inferred even without specific knowledge, based on everyday assumptions.  
The full placement of these objects is shown in Fig.~\ref{fig:object_placement}, and this classification is used in designing task instructions and experimental conditions.

To clarify the prior knowledge used in task allocation, we standardized the definition and labeling of objects commonly placed in everyday settings.  
Each object was evaluated using OpenAI’s GPT-4 (API model name: gpt-4) to estimate its most likely room among the 10 candidates shown in Fig.~\ref{fig:object_placement}.  
For object detection, we used YOLOv5~\cite{jocher2022ultralytics} to automatically extract 24-class object labels from environmental images.  
These labels are subsequently input to the LLM prompts for task decomposition and allocation, ensuring consistent integration of common-sense reasoning and on-site knowledge.

\begin{figure}[tb]
    \centering
    \includegraphics[width=0.7\linewidth]{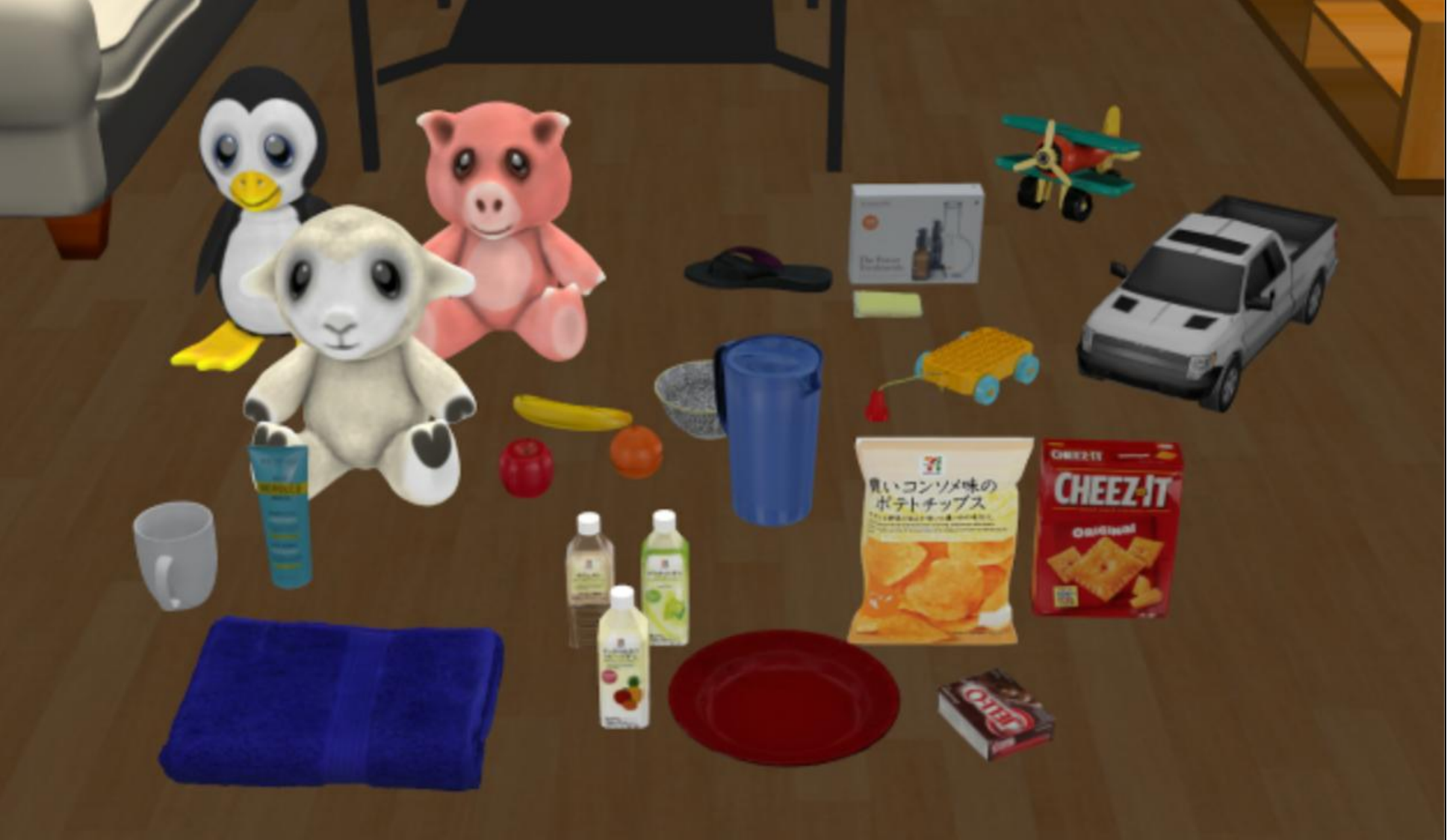}
    \caption{Set of 24 detectable objects (e.g., plate, bowl, banana, cracker\_box, pig\_doll, airplane\_toy, cup, sponge). Object labels obtained via YOLOv5 are used as input to LLM prompts and spatial concept learning.}
    \label{fig:objects}
\end{figure}

We empirically validated the task decomposition and robot assignment pipeline using object search tasks based on user instructions.  
As shown in Fig.~\ref{fig:objects}, target objects were specified by the user and mapped to likely locations based on spatial knowledge (room names and room-wise object presence probabilities).  
Subsequently, tasks were decomposed into action units (e.g., determining which room to navigate to and which object to detect/grasp), and each subtask was assigned to the robot with knowledge of the corresponding area.  
This pipeline enables high-level user instructions to be translated into concrete multi-robot action sequences for integrated evaluation of search, retrieval, and coordination.

To ensure replicability and consistency, the entire framework was implemented on the robotic development platform~\cite{el2022software}, using Toyota's Human Support Robot (HSR)~\cite{yamamoto2019development}. 
Each robot was assigned to a single floor (1F or 2F) and maintained an occupancy grid map specific to its designated area.  
Before execution, each robot learned the spatial concept model for its own floor, allowing all subsequent processes—task decomposition, allocation, and execution—to depend consistently on this acquired knowledge.


\subsection{Preprocessing}
\label{sec:Experiment_preparation}

Before the experiments, occupancy grid maps were built using Hector SLAM~\cite{kohlbrecher2011flexible}.  
HSRs were teleoperated to traverse each floor while acquiring range measurements via LiDAR sensors.  
This ensured accurate base maps for spatial concept learning, task decomposition, and allocation, and enabled stable self-localization via Monte Carlo Localization (MCL).

For object detection, training settings were standardized to strike a balance between reproducibility and performance.  
YOLOv5 was trained using 14,140 training images and 707 validation images for the 1F model, and 8,600 training images and 430 validation images for the 2F model.  
Training was performed for 30 epochs with batch size 8. Pretrained weights (\texttt{yolov5m}) were used\footnote{https://github.com/ultralytics/yolov5/blob/master/models/yolov5m.yaml}.  
Detected object labels were consistently supplied to both the spatial concept model and LLM prompts.

For spatial concept learning, the following hyperparameters were used for reproducibility and inference stability:  
$\alpha = 2.0$, $\gamma = 0.5$, $\beta = 0.1$, $\chi = 0.1$, $m_0 = [0,0]^\top$, $\kappa = 1.0$, $V_0 = \mathrm{diag}(2, 2)$, $\nu_0 = 3.0$, $\lambda = 0.1$.  
The number of particles was set to 30, and the fixed-lag window to 10 for approximation.

Each robot sequentially learned spatial concepts (including object observations) in the following order:

\begin{itemize}
  \item 1F: Entrance → Dining Room → Living Room → Office Room → Kitchen
  \item 2F: Front of Stairs → Corridor → Bathroom → Child's Room → Parent's Room
\end{itemize}

Each of the five rooms on each floor was visited 30 times (150 sessions per floor).  
During each session, one image, 2D self-position via MCL, object detection results, and location-related vocabulary were integrated into the observation used to update room-wise object presence probabilities by room.

To evaluate allocation robustness across difficulty levels, instruction types were categorized into four types based on the object list in Fig.~\ref{fig:objects}:

\begin{itemize}
  \item \textbf{Random}: Randomly select one object from each floor, regardless of placement type.
  \item \textbf{Hard-to-Predict}: Randomly select one hard-to-predict object per floor.
  \item \textbf{Common-sense}: Randomly select one common-sense-placed object per floor.
  \item \textbf{Mixed}: Select one hard-to-predict and one common-sense object, one from each floor.
\end{itemize}

For each task, GPT-4 generated an English instruction sentence (e.g., ``Could you please find apple.'' or ``I need you to locate penguin doll.'') based on the YOLOv5-detected object label.  
Synonyms such as ``find'' and ``search for'' were mixed to ensure diversity in prompt phrasing.  
These instructions were input to the task decomposition module to evaluate allocation performance consistently.

\subsection{Baselines}
\label{sec:comparative approach}

To quantify the allocation effectiveness of our method, we compared it with the following two baselines that do not use spatial concept models:

\begin{itemize}
  \item \textbf{Random Allocation}: Subtasks generated from decomposition are randomly assigned to one of the two robots without referencing any knowledge. This serves as a lower bound.
  \item \textbf{GPT-4 Commonsense Allocation}: Allocation is based solely on GPT-4's commonsense reasoning about object location, without using learned spatial concepts.
\end{itemize}

These comparisons allow us to isolate the contribution of:  
(i) the use of any allocation policy versus random (knowledge-free), and  
(ii) the added value of spatial concepts and room-wise object presence probabilities over commonsense-based inference alone.

\subsection{Evaluation Metrics}
\label{sec:evaluation item}

The evaluation metric is the {Task allocation success counts}.  
A subtask is considered successful if the robot to which it is assigned has, based on prior learning, acquired on-site knowledge (e.g., presence probability of the target object) corresponding to the target area.  
This criterion enables us to evaluate whether, even under the constraint of robots being confined to separate floors, the decomposed subtasks consistently reach the appropriate knowledge-holding robot.  
Thus, this metric effectively assesses whether the proposed allocation mechanism is practically viable under real-world operational constraints.

\subsection{Experimental Results}
\label{sec:experimental results}

    \begin{table}[tb]
      \centering
      \caption{room-wise object presence probabilities by location (Robot1)}
      \label{tab:robot1_probs}
      \begin{adjustbox}{max width=\linewidth}
      \begin{tabular}{l *{5}{S}}
        \toprule
        Object & {entrance} & {dining} & {living\_room} & {office\_room} & {kitchen}\\
        \midrule
        pitcher\_base & 0.136 & \bfseries 0.848 & 0.004 & 0.006 & 0.006 \\
        bowl          & 0.136 & \bfseries 0.848 & 0.004 & 0.006 & 0.006 \\
        plate         & 0.309 & 0.010 & \bfseries 0.662 & 0.010 & 0.009 \\
        coffee        & 0.152 & 0.006 & 0.005 & \bfseries 0.831 & 0.006 \\
        towel         & 0.120 & 0.006 & 0.056 & \bfseries 0.813 & 0.006 \\
        penguin\_doll & 0.271 & 0.009 & \bfseries 0.702 & 0.009 & 0.009 \\
        sheep\_doll   & 0.278 & 0.009 & \bfseries 0.694 & 0.010 & 0.009 \\
        pudding\_box  & 0.278 & 0.009 & \bfseries 0.694 & 0.010 & 0.009 \\
        fruits\_juice & 0.250 & 0.009 & \bfseries 0.668 & 0.066 & 0.008 \\
        tooth\_paste  & 0.328 & 0.010 & 0.006 & 0.011 & \bfseries 0.645 \\
        apple         & 0.248 & 0.008 & 0.005 & 0.009 & \bfseries 0.729 \\
        orange        & 0.200 & 0.007 & 0.005 & 0.008 & \bfseries 0.779 \\
        muscat        & 0.200 & 0.007 & 0.005 & 0.008 & \bfseries 0.779 \\
        \bottomrule
      \end{tabular}
      \end{adjustbox}
      \vspace{2pt}
      \footnotesize\emph{Bold = most likely location per object.}
    \end{table}
    
    \begin{table}[tb]
      \centering
      \caption{room-wise object presence probabilities by location (Robot2)}
      \label{tab:robot2_probs}
      \begin{adjustbox}{max width=\linewidth}
      \begin{tabular}{l *{5}{S}}
        \toprule
        Object & {front\_of\_stairs} & {corridor} & {bathroom} & {child\_room} & {parent\_room}\\
        \midrule
        car\_toy       & \bfseries 0.899 & 0.087 & 0.004 & 0.005 & 0.004 \\
        airplane\_toy  & 0.011 & 0.223 & \bfseries 0.753 & 0.007 & 0.006 \\
        body\_sponge   & 0.011 & 0.223 & \bfseries 0.753 & 0.007 & 0.006 \\
        bath\_slipper  & 0.011 & 0.223 & \bfseries 0.753 & 0.007 & 0.006 \\
        truck\_toy     & 0.012 & 0.264 & 0.006 & \bfseries 0.711 & 0.006 \\
        pig\_doll      & 0.012 & 0.264 & 0.006 & \bfseries 0.711 & 0.006 \\
        cracker\_box   & 0.012 & 0.264 & 0.006 & \bfseries 0.711 & 0.006 \\
        chips\_bag     & 0.012 & 0.264 & 0.006 & \bfseries 0.711 & 0.006 \\
        cup            & 0.011 & 0.213 & 0.006 & 0.007 & \bfseries 0.764 \\
        banana         & 0.010 & 0.186 & 0.130 & 0.007 & \bfseries 0.668 \\
        treatments     & 0.011 & 0.213 & 0.006 & 0.007 & \bfseries 0.764 \\
        \bottomrule
      \end{tabular}
      \end{adjustbox}
      \vspace{2pt}
      \footnotesize\emph{Bold = most likely location per object.}
    \end{table}


\begin{table*}[tb]
    \centering
    \caption{Task allocation success counts across instruction types}
    \label{tbl:result}
    \begin{tabularx}{\textwidth}{lXXXXc}
        \toprule
        \diagbox{Method}{Instruction Type} 
            & Random
            & Hard-to-Predict
            & Common-Sense
            & Mixed
            & Total \\
        \midrule
        Proposed Method 
            & \textbf{17/20} 
            & \textbf{10/10} 
            & \textbf{10/10} 
            & \textbf{10/10} 
            & \textbf{47/50} \\
        Random Allocation 
            & \uline{11/20} 
            & \uline{6/10} 
            & 4/10 
            & \uline{7/10} 
            & \uline{28/50} \\
        GPT-4 Commonsense Allocation 
            & 10/20 
            & 3/10 
            & \uline{6/10} 
            & \uline{7/10} 
            & 26/50 \\
        \bottomrule
    \end{tabularx}
\end{table*}

Tables~\ref{tab:robot1_probs} and~\ref{tab:robot2_probs} list, for each object, the normalized probabilities of occurrence across rooms (normalized row-wise), with the most likely room highlighted in bold.  
This representation helps explain the effectiveness of initial room visits in qualitative rollout and the suppression of redundant searches, providing a rationale for the superior performance of the proposed method over random and commonsense-based baselines.

The comparative results in Table~\ref{tbl:result} show that the proposed method outperforms both baselines across all instruction categories.  
Our method achieved a success count of {47 out of 50}, significantly surpassing both {Random Allocation} ({28/50}) and {GPT-4 Commonsense Allocation} ({26/50}).  
In detail, the proposed approach succeeded in {17/20} cases under the ``Random per Floor'' condition, and achieved a perfect score of {10/10} in all three combined-task conditions: Hard-to-Predict, Common-Sense, and Mixed.  
This consistent advantage suggests that explicitly incorporating room-wise object presence probabilities—learned through spatial concept models—into the prompt design contributed directly to accurate subtask-to-robot assignments.  
Moreover, even under the constraint that robots are restricted to a specific floor, knowledge-based allocation consistently yielded highly reliable results.

A detailed analysis of failure cases reveals the conditions under which misallocations tend to occur.  
All observed failures occurred exclusively in the {Random per Floor} condition with single-object targets, specifically for ``car toy'', ``treatments'', and ``sponge''.  
These results indicate that when a non-qualified robot is included in the candidate set, prioritization based on existence probability weakens, leading GPT-4's inference to become more susceptible to error.  
In other words, when the subtask involves a single object and the knowledge gap between robots is small, the language model is more likely to be misled by general knowledge and incorrectly assign the task to an inappropriate robot.

Conversely, the behavior of the {GPT-4 Commonsense Allocation} method reveals notable vulnerabilities when relying solely on commonsense.  
Its success count under ``Random per Floor'' was slightly lower than that of the Random baseline, and more critically, in the ``Hard-to-Predict'' condition, it underperformed Random Allocation by 3 cases.  
This suggests that for objects whose placements are specific to the environment, LLM commonsense introduces a bias toward \emph{typical room associations}, increasing the probability of misallocating to a robot on the wrong floor—particularly in our setup where inter-floor traversal is prohibited.  
While commonsense may work reasonably well for typically placed objects, ignoring environment-specific presence distributions leads to inconsistent allocation decisions in disjointed environments.  
Therefore, our proposed strategy—embedding spatial concept–based existence probabilities into the prompt—proves essential in calibrating LLMs' commonsense bias and systematically preventing misallocations.

In summary, task decomposition and allocation based on spatial concepts consistently achieved high success rates, even in a two-zone environment where cross-floor movement is restricted.  
This confirms the effectiveness of calibrating GPT-4's commonsense inference using environment-specific spatial knowledge.  
At the same time, the observed vulnerability under single-object scenarios—where candidate robots include ineligible agents—highlights a potential weakness of purely language-driven reasoning.

\section{Experiment~2}
\label{chap:Experiment II}

\begin{figure}[tb]
        \centering
        \includegraphics[width=1\linewidth]{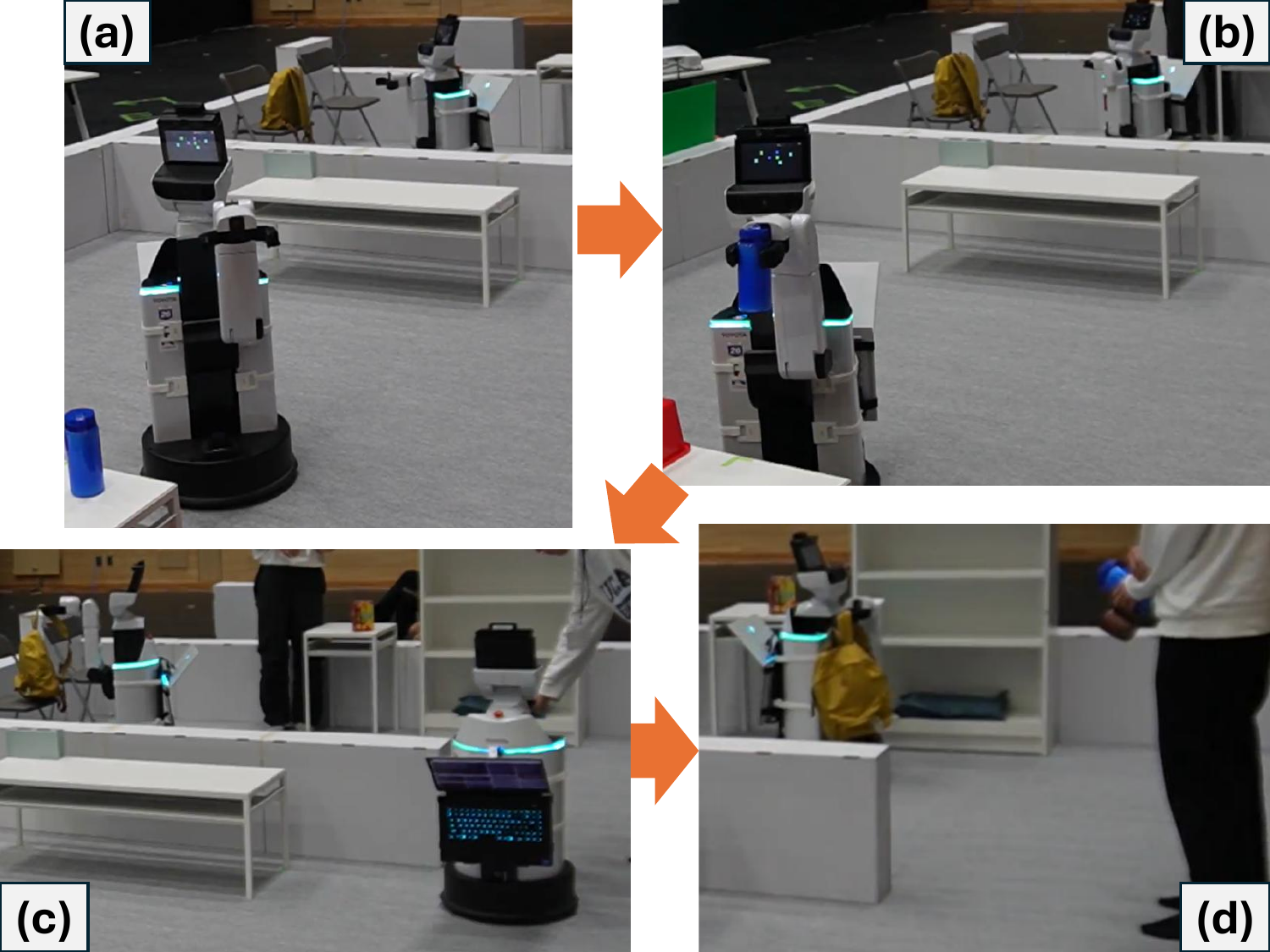}
        \caption{Real-world demonstration of the proposed system in the RoboCup @Home DSPL field.
        (a) Robot2 detects the location of the \textbf{water bottle}. 
        (b) Robot2 grasps the water bottle. 
        (c) Robot2 delivers the water bottle to the user, while Robot1 in the background grasps the \textbf{bag}. 
        (d) Robot1 halts during transportation due to interference between the grasped bag and its front safety sensors.
        }
        \label{fig:RoboCup_Image}
    \end{figure}

This section reports the qualitative validation of the proposed framework conducted during the RoboCup competition.  
The objective is to demonstrate the feasibility of our method through real-world deployment, as evidenced by Fig.~\ref{fig:RoboCup_Image} and supplementary demonstration videos recorded in both the RoboCup venue and our laboratory.

In this experiment, we deployed our proposed framework, which includes task decomposition, subtask allocation, sequential behavior planning, and execution, in a real-world household-like environment provided by the RoboCup @Home DSPL competition.  
The goal was to qualitatively assess its feasibility and robustness under real-world constraints and environmental noise.

\subsection{Experimental Setup}
\label{sec:experimental conditions}

\begin{figure}[tb]
    \centering
    \includegraphics[width=1\linewidth]{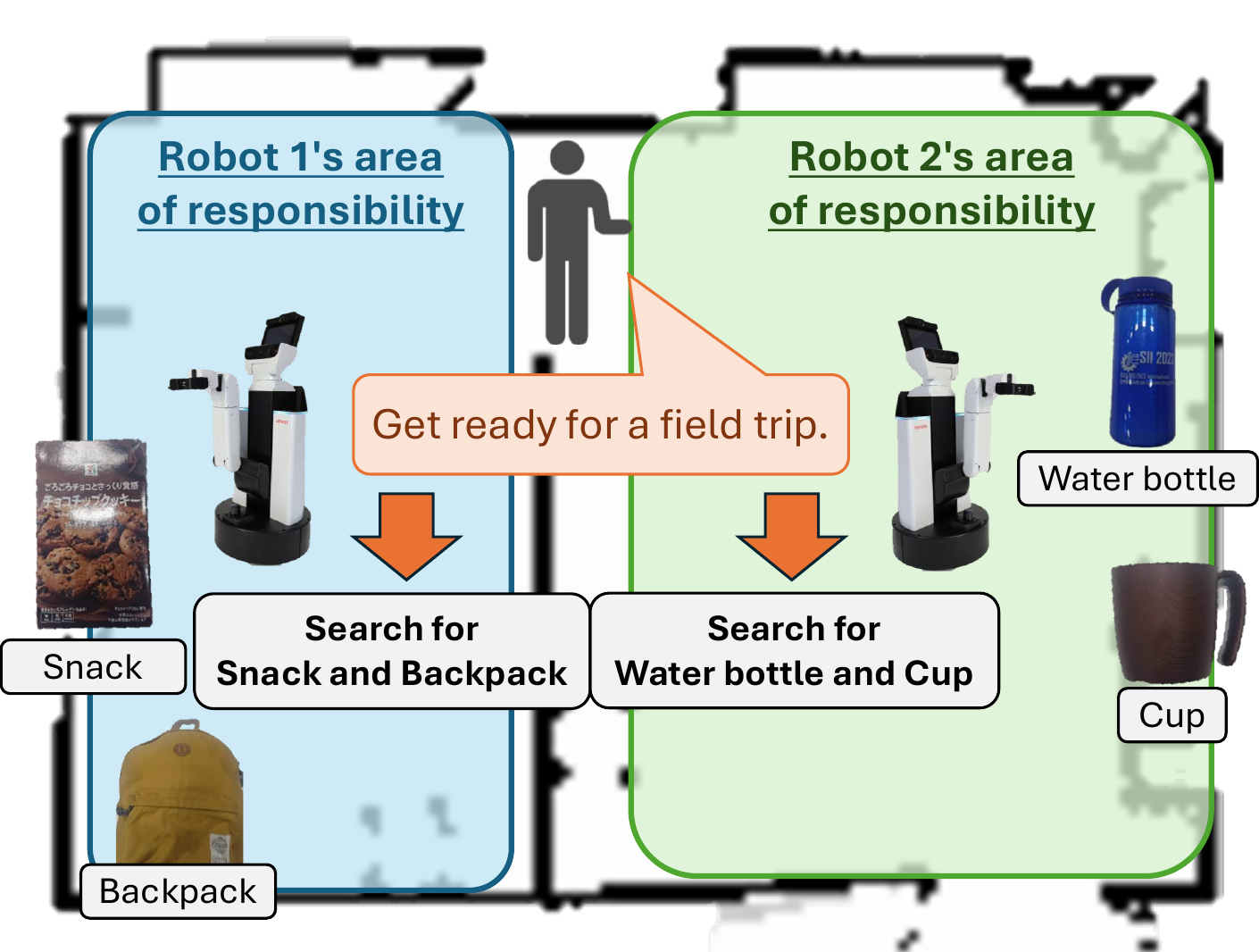}
    \caption{Top-down layout of the RoboCup @Home DSPL domestic environment. Solid black lines represent walls and fixed obstacles. Each labeled area corresponds to a ``room category'' in our framework. The room-wise object presence probabilities inferred from these categories were embedded into LLM prompts and used as grounding for subtask allocation.}
    \label{fig:RoboCup}
\end{figure}






\begin{table}[t]
\centering
\caption{Robot 2 action sequence (actual execution trace). Steps are chronological.}
\label{table:robot2_actions}
\begin{tabularx}{0.95\linewidth}{@{}r l X@{}}
\toprule
\textbf{Step} & \textbf{Skill} & \textbf{Argument} \\
\midrule
1  & navigation        & kitchen \\
2  & object\_detection & cup \\
3  & pick              & cup \\
4  & navigation        & gather \\
5  & place             & gather \\
6  & navigation        & kitchen \\
7  & object\_detection & water\_bottle \\
8  & pick              & water\_bottle \\
9  & navigation        & gather \\
10 & place             & gather \\
11 & navigation        & kitchen \\
\bottomrule
\end{tabularx}
\end{table}

The purpose of this real-world experiment was to validate the full execution flow—from interpreting an ambiguous instruction to physical task completion—for the input command:  
{``Get ready for a field trip.''}.  
The experiment was conducted under ambient noise and real-world environmental conditions in the RoboCup field (Fig.~\ref{fig:RoboCup}).  
Two HSR robots were used, each responsible for a specific zone within the field.

Each robot had pre-trained multimodal spatial concept models that allowed it to infer both {room categories} and {room-wise object presence probabilities} for its assigned area.  
This on-site knowledge was embedded into {prompts for GPT-4}, which interpreted the ambiguous instruction, decomposed it into subtasks, and assigned them to robots based on their spatial knowledge.

In this scenario, the LLM assigned the following subtasks:
\begin{itemize}
    \item \textbf{Robot1}: Retrieve a \texttt{bag} and some \texttt{snacks}
    \item \textbf{Robot2}: Retrieve a \texttt{water bottle} and \texttt{fruit juice}
\end{itemize}

Each robot then performed \texttt{navigation}, \texttt{object\_detection}, \texttt{pick}, and \texttt{place} actions in a sequentially planned manner.  
If a step failed, the robot autonomously reattempted the sequence within the scope of its current subtask.

The validation was carried out in the competition venue, where real-world challenges such as ambient noise and partial observability were present.

As an example of concrete skill execution, {Robot2} navigated back and forth between the \texttt{kitchen} and the \texttt{gathering area}, executing a series of \texttt{object\_detection}–\texttt{pick}–\texttt{place} actions.  
Table~\ref{table:robot2_actions} presents the actual execution trace recorded during this process.






\subsection{Result}
\label{sec:Qualitative results of the demonstration}    

    The LLM appropriately decomposed the ambiguous instruction ``Get ready for a field trip.'' into concrete subtasks, assigning ``bag'' and ``snacks'' to Robot1, and ``water bottle'' (and other drink-related items) to Robot2.  
Fig.~\ref{fig:RoboCup} visualizes the entire execution pipeline, from instruction input to final task execution, in the actual RoboCup venue.

In this scenario, the two HSRs utilized the room categories and room-wise object presence probabilities obtained from their spatial concept models to interpret the high-level instruction, perform task decomposition, allocate subtasks, plan sequential actions, and execute physical tasks.  
This was achieved in a realistic setting with ambient noise, dynamic obstacles (e.g., human movement), and partial observability, within the simulated domestic field.

{Robot2 successfully detected and picked up a \texttt{cup} in the \texttt{kitchen}, delivered it to the \texttt{gather} area, and then completed a similar action sequence for the \texttt{water\_bottle}}.  
Table~\ref{table:robot2_actions} shows the actual execution log, illustrating how the LLM-generated subtask sequence was translated into concrete robot actions and manifested as physical round trips in the environment.

On the other hand, {Robot1 halted during the delivery phase after successfully picking up the ``bag''}.  
The cause was that the pose of the gripped object interfered with the front safety sensors, triggering a persistent emergency stop, from which the robot could not recover autonomously.  
Although re-planning was attempted, the issue remained unresolved.  
In this demonstration, we treated this malfunction as an exception.  
While task decomposition, allocation, and partial execution (by Robot2) were successful, the incident highlighted the need to explicitly account for grasping pose constraints and collision avoidance during planning.

Due to privacy concerns, the recorded RoboCup footage is not publicly available.  
Instead, we provide a supplementary laboratory video that shows a successful run under the same framework and experimental settings as Experiment~II.\footnote{https://youtu.be/LMoWAp\_kPhk}

Although the OIT--RITS joint team~\footnote{https://www.youtube.com/@oit-rits/featured} (which implemented this system) placed among the top teams in the Open Challenge at the competition, the results presented here are not intended as a quantitative evaluation of competition performance.  
Rather, the purpose of this section is to provide a qualitative demonstration of feasibility in a real-world deployment context.





\section{Limitation}
\label{sec:limitation}
This section outlines three key limitations of our framework: (i) limited support for tasks requiring synchronous cooperation, (ii) limited robustness to execution-time failures and the absence of dynamic reallocation, and (iii) the homogeneous-team assumption (two identical HSRs).

The primary limitations of the proposed framework are its vulnerability to tasks that require synchronous cooperation and its lack of robustness in the face of failures during execution.  
Specifically, simultaneous collaborative behaviors—such as dual-arm manipulation, object handover, and cooperative transportation—are difficult to manage under the current assumption of independent subtasks.  
Moreover, if one robot becomes non-operational, subtask reassignment does not occur automatically, which may result in incomplete task execution.  
To address these issues, the framework must be extended to support dynamic task reallocation based on runtime observations and failure signals, as well as synchronized planning with inter-robot communication and shared context~\cite{liu2024coherent,yang2025autohma}.

Another critical limitation stems from the assumption that all robots are of the same type and capability.  
In this study, two identical HSRs were used, each assumed to be capable of executing subtasks independently.  
However, deploying multiple high-cost robots with similar performance in real domestic environments is often impractical due to economic and logistical constraints.  
Future work should consider heterogeneous robot teams~\cite{liu2024coherent,yang2025autohma,el2025public}, including mobile bases, stationary manipulators, and lightweight platforms, by explicitly modeling differences in capabilities such as payload, reachable workspace, and gripper types.  
This would enable a skill- and cost-aware task allocation mechanism that accounts for functional disparities and resource constraints, improving the feasibility of real-world deployment.

Finally, the current evaluation primarily focuses on the number of subtasks assigned correctly to the appropriate robot, thereby emphasizing assignment accuracy.  
As a result, other operational factors—such as long-distance navigation, task duration, energy consumption, and behavior under crowded or dynamic conditions—were not comprehensively evaluated.  
A more systematic analysis of runtime performance is necessary to fully assess the practicality and efficiency of the proposed framework in real-world scenarios.

\section{Conclusion}
\label{chap:conclusion}
\subsection{Summary}
\label{sec:conclusion}

This study proposed a framework that integrates an LLM with a spatial concept model for multi-robot task decomposition and knowledge-based assignment.  
By leveraging room names and room-wise object presence probabilities inferred from the spatial concept model, the system decomposes user instructions into subtasks and allocates them to robots more accurately than baseline methods.  
We extended SMART-LLM, a multi-robot task assignment method, by incorporating environment-specific knowledge from each robot and enabling the LLM to infer which robot is more likely to complete each subtask.  

In simulation experiments, task allocation based on learned room names and room-wise object presence probabilities consistently achieved higher success rates than baselines across all task types, validating the effectiveness of our method.
Furthermore, the feasibility of the proposed approach was confirmed through a real-world demonstration in the RoboCup @Home competition. 
These results show that our method is not only effective in simulation but also applicable to real environments, supporting its potential for deployment in household service robots.








\subsection{Future Work}
\label{sec:future}

We plan to incorporate skill-aware allocation mechanisms that explicitly model inter-robot capability differences (e.g., payload capacity, reachability, grasping precision) and assign subtasks accordingly, enabling a quantitative assessment of their impact on performance and robustness.

We plan to improve system reliability by introducing runtime monitoring and dynamic reallocation in response to incorrect or failed assignments, and by generalizing the framework to heterogeneous robot teams. This includes integrating progress monitors and safety preemption, and evaluating cooperation under realistic uncertainties such as asymmetric sensing, communication delays, and physical disparities.

In the longer term, we plan to extend the framework to human–robot collaboration and continual learning guided by uncertainty estimation, incorporating confirmation requests or minimal human intervention based on the confidence of LLM outputs. We also aim to establish a benchmark platform that enables unified evaluation of decomposition, allocation, and execution in realistic environments.






\section*{Acknowledgments}
This work was partially supported by the JST Moonshot R\&D Program (Grant Number JPMJMS2011), JSPS KAKENHI (Grant Numbers JP25K15292 and JP23K16975), and the JST Challenging Research Program for Next-Generation Researchers (Grant Number JPMJSP2101).


\bibliographystyle{spmpsci_change}
\bibliography{refs}

\end{document}